\documentclass{ecai} 

\usepackage{latexsym}
\usepackage{amssymb}
\usepackage{amsmath}
\usepackage{amsthm}
\usepackage{booktabs}
\usepackage{enumitem}
\usepackage{graphicx}
\usepackage{color}

\newcommand{\BibTeX}{B\kern-.05em{\sc i\kern-.025em b}\kern-.08em\TeX}

\begin{document}


\begin{frontmatter}


\paperid{2338} 


\title{Estimating Player Performance in Different Contexts Using Fine-tuned Large Events Models}


\author[A,B]{\fnms{Tiago}~\snm{Mendes-Neves}\thanks{Corresponding Author. Email: tiago.neves@fe.up.pt}} 
\author[B]{\fnms{Luís}~\snm{Meireles}} 
\author[A,C]{\fnms{João}~\snm{Mendes-Moreira}} 

\address[A]{Faculdade de Engenharia Universidade do Porto}
\address[B]{Nordensa Football}
\address[C]{LIAAD - INESC TEC}


\begin{abstract}

This paper introduces an innovative application of Large Event Models (LEMs), akin to Large Language Models, to the domain of soccer analytics. By learning the "language" of soccer — predicting variables for subsequent events rather than words — LEMs facilitate the simulation of matches and offer various applications, including player performance prediction across different team contexts. We focus on fine-tuning LEMs with the WyScout dataset for the 2017-2018 Premier League season to derive specific insights into player contributions and team strategies. Our methodology involves adapting these models to reflect the nuanced dynamics of soccer, enabling the evaluation of hypothetical transfers.
Our findings confirm the effectiveness and limitations of LEMs in soccer analytics, highlighting the model's capability to forecast teams' expected standings and explore high-profile scenarios, such as the potential effects of transferring Cristiano Ronaldo or Lionel Messi to different teams in the Premier League. This analysis underscores the importance of context in evaluating player quality. While general metrics may suggest significant differences between players, contextual analyses reveal narrower gaps in performance within specific team frameworks.

\end{abstract}

\end{frontmatter}

\section{Introduction}

Effectively navigating decision-making in the sports domain, especially when substantial financial investments in players or managers are at stake, presents a complex challenge. Despite the substantial growth in using data to improve decision-making in recent decades \cite{statsbomb_what_nodate}, its application often needs a comprehensive approach. An illustrative example highlighting this need is the assessment of compatibility between a specific player and a particular team. While various companies in the market offer "AI-based solutions" for determining this compatibility, there is a notable absence of published validation studies for such solutions. Many of these solutions may be grounded in a linear paradigm, whereas the complexity of the problem clearly suggests the requirement for a non-linear method.

In the meantime, roughly 50\% of soccer transfers fail \cite{tom_worville_how_2021}, which can be linked to diverse factors (see Table \ref{tab:reasons_for_failure}). Certain aspects, such as adaptability issues, might be beyond the control of sporting directors or other decision-makers. We assert that it is possible to quantify whether a specific player aligns with a particular playing style and if his integration into a team ultimately proves beneficial, but this cannot be achieved through traditional methods.

\begin{table}[]
    \centering
    \begin{tabular}{c}
    \hline
        Player not as good as thought. \\
        Player doesn’t fit style.*\\
        Played out of position.*\\
        Manager doesn’t like.\\
        Fitness/personal issues.\\
        Current player is better.*\\
    \hline
    \end{tabular}
    \caption{Why transfers fail according to Ian Graham, compiled by Tom Worville \cite{tom_worville_how_2021}. *Reasons that can be addressed in our framework.}
    \label{tab:reasons_for_failure}
\end{table}

In a broad sense, this study aims to evaluate a player's impact on a soccer team, thus aligning with various existing studies. Numerous methods employ frameworks that assess events based on game state value, commonly called possession value \cite{statsbomb_what_nodate}. One example is the Valuing Actions by Estimating Probabilities (VAEP) model, which treats the estimation of game state value as a machine learning challenge \cite{decroos_actions_2019}. It calculates the probability of an action leading to a goal within subsequent actions. Another approach involves the expected threat (xT) metric, assigning values to different areas of the pitch based on the likelihood of scoring from those zones \cite{statsbomb_what_nodate}. These methods go beyond event data, providing solutions for tracking data \cite{fernandez_framework_2021}. Furthermore, we can also see some relationship between our approach and models designed to estimate player transfer fees \cite{mchale_estimating_2023}, valuing players over extended periods \cite{mendesneves2022valuing}, and other related aspects. The significance of these methods cannot be overstated. Research indicates that following the Moneyball paradigm \cite{lewis_moneyball_2004}, most teams rapidly adopted this data-centric approach in their recruitment strategies \cite{hakes_economic}.

However, the current approaches neglect to account for the impact of contextual factors on the values they measure. This gap becomes evident when addressing a fundamental question: How would these values change if a player were to join a different team? Despite notable research on how VAEP varies with a player's league \cite{the_come_on_man_predicting_nodate} and the influence of players on action outcomes \cite{tahmeed_tureen_estimated_2022}, this particular question remains largely unanswered.

Contrasting with conventional methodologies, we use Large Events Models (LEMs) [12]. LEMs operate on principles similar to Large Language Models (LLMs), predicting the next element in a sequence by considering the existing context\footnotemark[1]. In LLMs, the subsequent element is a word influenced by preceding words. Conversely, the next element in LEMs is an event shaped by the current game state. The upcoming element can alter the present context in LEMs and LLMs, enabling the models to generate large amounts of coherent information from a given start point.

\footnote{Context can refer to two different concepts over this paper: (1) context is the input given to LEMs and LLMs that informs the models on how to forecast the next token, and (2) context as the environment of a club, its players and playstyle, which affects well a player performs on a team}

This methodology addresses a specific issue: while LLMs excel in creative tasks, they struggle in reasoning tasks \cite{huang2023reasoning}. For example, LLMs face challenges in generating a concise list of players who could improve a team. LEMs, on the other hand, enable the simulation of diverse contexts and the potential behavior of a player within those contexts. This capability facilitates informed decision-making based on multiple Key Performance Indicators (KPIs). These include anticipated contributions to the team in terms of points, shots, key passes, crosses, and set-piece goals.

To answer these questions, we fine-tune LEMs to learn team and player behaviors across different scenarios. Our analysis involves assessing teams from the 2017/18 English Premier League (EPL) season, demonstrating LEMs' capacity to capture distinct soccer team behaviors. Additionally, we conduct a comparative analysis for each Premier League team, evaluating the potential impact of acquiring either Messi or Ronaldo. This comparison helps clarify our methodology's strengths and limitations, providing insights into its accuracy and areas of improvement.

While the availability of public data constrains our study, the methodology we propose enhances the predictive depth beyond current literature offerings. By leveraging LEMs, our approach facilitates a detailed analysis of a player's potential impact when joining a new team. This method outperforms the numerical summation of player attributes, comprehensively exploring the outcomes of recruiting a new player. The impact is assessable across a broad spectrum of metrics, enriching knowledge and enhancing the effectiveness of recruitment decisions.

The paper is structured as follows:
\begin{itemize}
    \item 
Section \ref{sec:background} provides the background knowledge on LEMs.
    \item 
Section \ref{sec:methodology} describes the methodology used in this work, including the data, parameter tuning, metrics, and limitations.
    \item 
Section \ref{sec:experiments} presents and discusses the experiments performed.
    \item 
Section \ref{sec:discussion} discusses the insights obtained in this work.
    \item 
Section \ref{sec:conclusion} presents the concluding remarks of this work.
\end{itemize}

\section{Background}\label{sec:background}
\subsection{Large Events Models}

\begin{figure}[h]
  \centering
  \includegraphics[width=\linewidth]{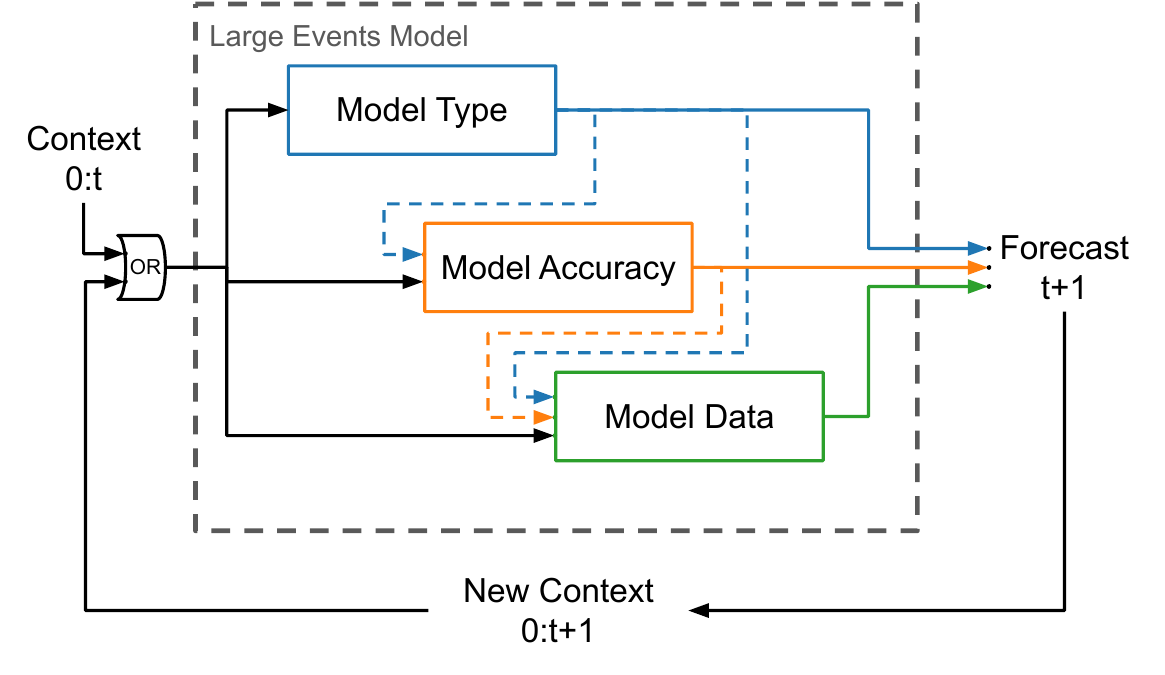}
  \caption{LEMs and LLMs work on the same principle: given the initial context that contains all current information, it can forecast the next token. This token then updates the context iteratively until an exit criterion is met. For LLMs, the tokens are words. For LEMs, the tokens are events.}
  \label{fig:fig1}
\end{figure}

LEMs draw inspiration from LLMs, applying their sequential prediction approach to the domain of soccer \cite{mendes_neves_towards_LEM}. LLMs utilize the context provided by preceding words to forecast the subsequent word in a text, with each new word modifying the context for continuous, coherent text generation. Similarly, LEMs utilize the current state of a soccer game to anticipate the next event. This model aims to predict the forthcoming event in a series based on the current game state (i.e., current result, previous events). Leveraging the sequential nature of soccer event data, LEMs enable large-scale simulation of soccer matches. These simulations are grounded entirely in data, with the model perpetually updating the game state with each new predicted event, as depicted in Figure \ref{fig:fig1}.

This approach provides many advantages over traditional machine learning approaches. It can be used to infer basic statistical approaches like expected goals \cite{Pollard_2004} or to simulate large amounts of data \cite{mendes_neves_datadriven_simulator_2021}. Most relevant is that it offers a framework that uses the same backbone to generate many insights through simulation. Even smaller LEMs can be used to simulate a restricted version of the soccer match, focusing only on key events \cite{simpson_seq2event_2022,yeung_transformer-based_2023}.


Our LEM uses a phased approach to ensure that the multiple variables we need to forecast are aligned. For example, if the following action is a pass, then the likelihood of being an accurate action is much higher when compared to being a shot. If we tried to forecast the two variables simultaneously, the error would be much higher than forecasting them sequentially. Therefore, the LEM sequentially forecasts the variables in the following order:
\begin{enumerate}
    \item 
\textbf{Model Type} forecasts the next event type.
    \item 
\textbf{Model Accuracy} forecasts the accuracy of the next event and if the next event is a goal.
    \item 
\textbf{Model Data} forecasts the next event location, the time elapsed until the next event, and if the home team performs the next event.
\end{enumerate}

The LEM forecasts the probability distribution for each variable. For example, the output of Model Type is the probability for each possible event. In a simulation environment, and akin to LLMs, we then sample the distribution using the aforementioned probabilities to obtain the next event type.
This means every forecast is probabilistic, which means that, for the same context, the output will vary according to the forecasted probabilities.

The features utilized in our LEMs are the following:
\begin{itemize}
    \item \textbf{Event Type}: A one-hot encoded categorical feature representing the type of event, such as passes, shots, or fouls, with 33 distinct types considered individually due to their unique distributions related to game context (e.g., location, score).
    \item \textbf{Period}: A binary variable indicating the match period (first half = 0, second half = 1), excluding extra time or penalties.
    \item \textbf{Minute}: A normalized continuous variable indicating the time of the event occurrence, adjusted to a 0-1 scale by dividing by 60.
    \item \textbf{X} and \textbf{Y}: These are continuous features representing the spatial coordinates on the soccer pitch, where the origin (0,0) is located at the bottom left corner from the perspective of the attacking team. The X-coordinate measures the horizontal location from one team's goal to another, while the Y-coordinate measures the vertical location. Both coordinates are normalized to a range from 0 to 1, ensuring a unified scale regardless of the actual pitch size. This system ensures that actions are consistently recorded from the perspective of the attacking team, thereby standardizing event locations across different matches.
    \item \textbf{IsHomeTeam}: A binary indicator denoting whether the event was performed by the home team (1) or the away team (0).
    \item \textbf{IsAccurate}: A binary indicator of whether the event was executed accurately (1) or not (0).
    \item \textbf{IsGoal}: A binary indicator set to 1 if the event resulted in a goal, otherwise 0.
    \item \textbf{HomeScore} and \textbf{AwayScore}: Normalized continuous variables representing the scores of the home and away teams at the event time, scaled to the interval [0,1] by dividing by 10.
\end{itemize}

The LEM forecasts the same variables as the inputs, with some exceptions: (1) the model forecasts the \textit{TimeElapsed} variable, which quantifies the temporal distance between consecutive events, rather than predicting \textit{Period} and \textit{Minute}, and (2) \textit{HomeScore} and \textit{AwayScore} are excluded from prediction, as these scores are directly influenced by the outcomes of other variables that the model forecasts (i.e. if an event leads to a goal, the score changes deterministically). 

To generate insights, LEMs are used as a simulation environment. Following the generation of predictions, an interpretation function updates the game state variables (i.e., the context). The new context is then used to generate the next event. This process can be repeatedly applied to simulate the sequence of events throughout a game.

LEMs have advantages and limitations. The positives include their foundational role in creating various metrics for soccer players, ranging from general performance in a season to specific impacts on aspects like shooting quality in particular zones. Simulating numerous games also allows for extracting patterns in the data, enabling comparison with baseline or affected patterns to create insightful metrics [15]. However, a significant limitation is that the model simulates generic football matches, as it was trained on data from all teams, resulting in average game simulations. In Section \ref{sec:methodology}, we will propose a methodology to enable context-specific insights by fine-tuning the models to replicate the behavior of specific teams and players to understand how changing a player’s context affects their performance.

Another large advantage of LEMs is that they can be built with any scope of data. There are no differences in methodology to build a LEM for women's or men's soccer, first tier versus non-league, among others. Furthermore, similar methods can be applied to tracking data. This significant advance leads us towards a complete, general model that can solve all problems in the game of soccer.

From a technical standpoint, LEMs use deep learning to learn the models. In theory, most machine learning algorithms can model these behaviors, with decision-tree-based ensembles performing the best. However, from a practical aspect, the models require fast inference times. This is where Deep Learning with GPUs comes in handy. The inference time is magnitudes faster than any other.

We used PyTorch to build and fine-tune the models to the specific data in this specific implementation. We provide the code to replicate all the aspects of this work, including the training of the base LEMs, at 
\url{https://github.com/nvsclub/LargeEventsModel}.

\section{Methodology}\label{sec:methodology}
\subsection{Data}

Event data in football refers to a structured record of specific occurrences during a match. These occurrences encompass a wide range of actions, including passes, shots, fouls, tackles, and goals. Each event is time-stamped and associated with various attributes such as player IDs, event types, locations on the field, and outcomes. This granular level of data provides a comprehensive view of the dynamics and flow of the game, allowing for in-depth analysis and insights into team strategies, player performance, and overall match dynamics.

This study relies on the Public Wyscout dataset \cite{pappalardo_public_2019} as the primary source of football event data. This dataset encompasses fixtures from top-tier leagues, including the Premier League (England), La Liga (Spanish), Serie A (Italy), Bundesliga (Germany), and Ligue 1 (France), with a focus on the 2017/2018 season. It provides an extensive range of events, from passes and shots to aerial duels and saves. We opted for this dataset since it was the most extensive available at the start of the project. Although the size of the dataset limits the comprehensiveness of the study, it was important for us to provide an open platform to incentive collaboration for further developments.

Our study used data from Ligue 1, encompassing 380 games, and Bundesliga, with 306 games, as the primary training set for the foundational Large Event Models (LEMs). For model validation, we employed data from 380 games in Serie A. To further refine and enhance the models, we incorporated data from the Premier League and La Liga (380 games each), using this additional information to extract detailed insights.

\subsection{Fine Tuning LEMs to Learn Different Contexts}

\begin{figure}[h]
  \centering
  \includegraphics[width=\linewidth]{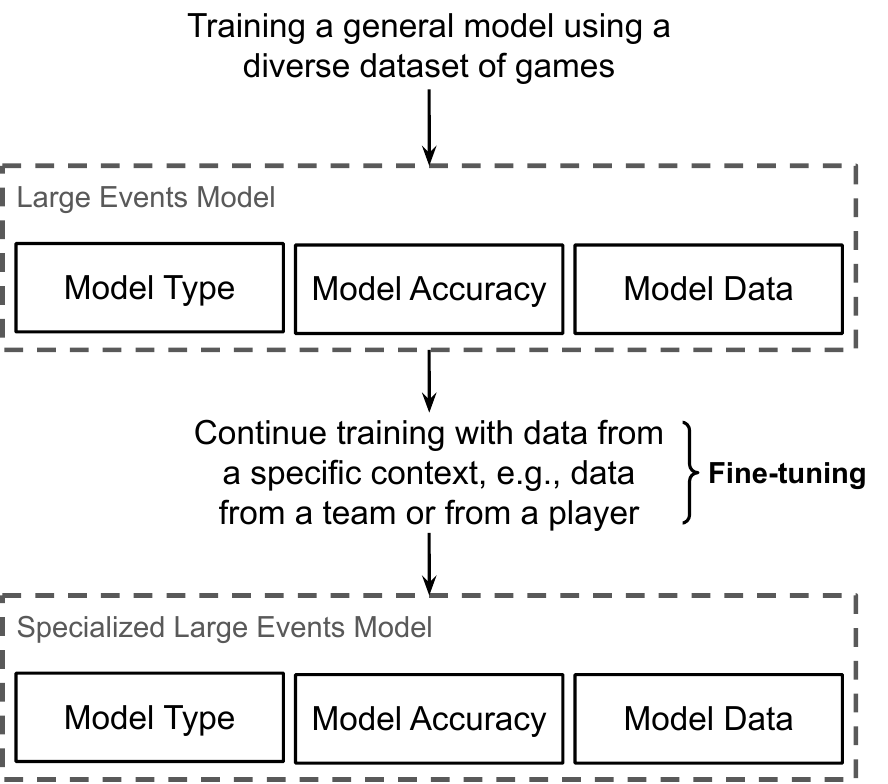}
  \caption{This figure depicts the two-stage process of developing a fine-tuned LEM model: first, we use a large dataset to build a LEM, then we fine-tune it using specific data about our target.}
  \label{fig:fig2}
\end{figure}

Fine-tuning means adjusting a pre-trained model on a specific task or dataset to improve performance. It involves further training a model already trained on a large dataset, typically on a more specialized or smaller dataset related to the task \cite{radford_improving_nodate}. The overall process can be seen in Figure \ref{fig:fig2}.

To learn new contexts, we fine-tune the models to reflect the specific behaviors of the teams. In essence, we change the general context-agnostic LEMs to perform another task. Instead of forecasting the next event in an average game of soccer, it will forecast the next event in a game within a predefined context. For this, we select the subset of data that is of interest. For example, if we want to build a model replicating the behavior of one team playing at home, we fine-tune the model using data from that specific team playing at home. Table \ref{tab:lem_fine_tune} presents the different types of contexts that we can learn in LEMs. For each type of fine-tuning, we reshape the training data to include exclusively some subsets of data, marked as "Includes". Sometimes overlap occurs. For example in player replacement, when including team data, it will necessarily include data from the player being replaced. Therefore, if we want to remove the effect of the player currently on the team that is going to be replaced by our target, we need to remove all data points related to the outgoing player.

\begin{table}[ht]
\centering
\begin{tabular}{ccccc}
\hline
\textbf{Fine-Tuning} & \textbf{Team} & \textbf{Opponent} & \textbf{Player} & \textbf{Player} \\
\textbf{Type} & \textbf{} & \textbf{} & \textbf{} & \textbf{Replaced} \\
\hline
Team & Includes & N/A & N/A & N/A \\
Player & N/A & N/A & Includes & N/A \\
Player Addition & Includes & N/A & Includes & N/A \\
Player Replacement & Includes & N/A & Includes & Removed \\
Team Face-off & Includes & Includes & N/A & N/A \\
\hline
\end{tabular}
\caption{The types of fine-tuning that we can perform on LEMs. The type "Team" focuses on replicating a single team's behavior against the league average. The type "Player" focuses on estimating the impact of the player on the average team of the league. "Player Addition" includes data from a new player in the context of the new team. "Player Replacement" does the same but excludes data from the player being replaced. In the "Team Face-off" type, we have the data of two teams: the home team and the away team.}
\label{tab:lem_fine_tune}
\end{table}

This paper focuses on the first four types of fine-tuned LEMs. We have excluded the "Team Face-off" model from this analysis. The rationale behind this exclusion is that "Team Face-off" primarily centers on evaluating potential tactical outcomes of two opposing teams rather than providing insights into player recruitment, which is the primary focus of this study.

\subsection{Player Addition vs Player Replacement}

Two categories can be used to measure a player's performance within a new context: "Player Addition" and "Player Replacement." The difference between both approaches is that Player Replacement actively removes the data from a competing player in the squad. On top of adding the influence of the incoming player, we remove the existing influence of the previous player on our squad. Therefore, instead of the training process slowly overwriting the data from the players, it is removed, allowing the training process to start from a blank slate. Table \ref{tab:player_replacements} shows the players chosen to be replaced. The criterion for selection of these players was "the attacking player with the least amount of game time from the top 11 most used players in the season".

\begin{table}[ht]
\centering
\begin{tabular}{cc}
\toprule
\textbf{Team} & \textbf{Player Replaced} \\
\hline
Man City & L. Sané \\
Liverpool & S. Mané \\
Tottenham & Son Heung-Min \\
Arsenal & A. Iwobi \\
Man United & R. Lukaku \\
Chelsea & V. Moses \\
Bournemouth & J. Ibe \\
Huddersfield & T. Ince \\
Newcastle & Joselu \\
Everton & D. Calvert-Lewin \\
Watford & A. Carrillo \\
Leicester & M. Albrighton \\
Southampton & N. Redmond \\
Brighton & S. March \\
Swansea & S. Clucas \\
WBA & J. Rodriguez \\
Burnley & A. Barnes \\
Stoke City & P. Crouch \\
Crystal Palace & C. Benteke \\
West Ham & M. Antonio \\
\toprule
\end{tabular}
\caption{The list of players that are replaced in the Player Replacement fine-tuning.}
\label{tab:player_replacements}
\end{table}

\subsection{Parameter Tuning}

We replicate the procedure to build the general LEM from Mendes-Neves et al. (2024) \cite{mendes_neves_towards_LEM}, described in Table \ref{tab:model_specs}.
\begin{table}[h]
\caption{Parameters used to train the general LEM.}
\label{tab:model_specs}
\begin{tabular}{@{}lllllll@{}}
\toprule
\textbf{Model} & \textbf{In} & \textbf{Out} & \textbf{Layers} & \textbf{L. Rate} & \textbf{Batch} & \textbf{Activ.} \\
\midrule
Type & 42 & 33 & [256] & 0.0010 & 32 & sigmoid \\
Accuracy & 75 & 2 & [128] & 0.0410 & 1024  & sigmoid  \\
Data & 77 & 264 & [64, 256, 256] & 0.0063 & 1024 & relu  \\
\toprule
\end{tabular}
\end{table}

Regarding the parameters used to fine-tune the models, we opted to maintain the original parameters used to train the general LEM, with the following exceptions:
\begin{itemize}
    \item 
The maximum number of epochs is 25.
    \item 
The learning rate lowered to 1/10th of the original value.
    \item 
The batch size follows Equation \ref{eq:batch}, where n is the number of events used for fine-tuning.
\end{itemize}

\begin{equation}
    \text{Batch Size} = \log(n^2), \quad \text{where} \quad 32 \leq \text{Batch Size} \leq 256
\label{eq:batch}
\end{equation}

However, we recommend optimizing these parameters through Bayesian optimization methods (same as the original LEM). We did not incur on this process due to our computational constraints.

\begin{figure}[h]
  \centering
  \includegraphics[width=\linewidth]{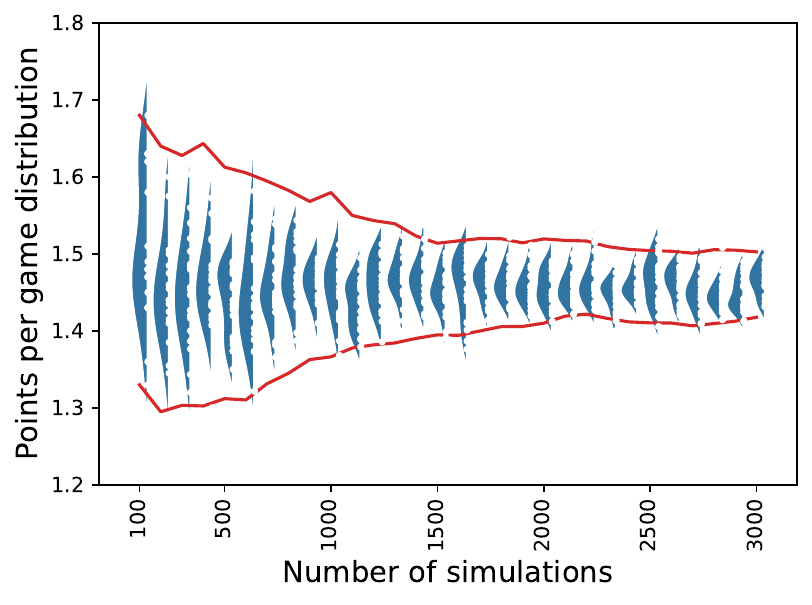}
  \caption{This figure illustrates that the outcome variability originates from the model's training and fine-tuning rather than the simulation process. At the threshold of 3000 simulations, a sequence of 10 consecutive wins is required to alter the expected points by a mere +0.01. Additionally, the figure shows that the error distribution follows a normal curve, indicating that, over an extended period, the average error in simulations is expected to converge towards zero.}
  \label{fig:variablility_analysis}
\end{figure}

The definition of the batch size is the most critical parameter in this work. Fine-tuning a model requires finding the right balance between frequent updates through backpropagation and the training speed of the models \cite{radford_improving_nodate}. Iteratively, we found Equation \ref{eq:batch} to provide the right balance for our use case. The other relevant parameter we must set is the number of simulations for each LEM. As visible in Figure \ref{fig:variablility_analysis}, the larger the number of simulations, the lower the variance of the results. We opted for a value of 2500 simulations since it finds a reasonable compromise between simulation time and low variance in the results. Furthermore, we train 10 iterations for each model, allowing us to evaluate the variability of the training process.

\subsection{Limitations}
This work is performed only for home games, but the work is extendable for away games. Since the methodology for away games is a mirror image of the process used to generate the results for home games, we opted to focus our analysis on the data from the home side. The results for the away perspective can be replicated by changing the flag “is\_home” in the test setup to “False.”

\section{Experiments}\label{sec:experiments}
Our framework is difficult to evaluate because we do not have an objective ground truth regarding the replication of teams through simulation. Therefore, our experiments aim to provide intuition into how our framework provides insights.

\subsection{Simulating the Premier League}

Table \ref{tab:football_positions} showcases the comparison of forecasts from the finetuned LEM models against the actual end-of-season results for the league. The table displays the expected positions as predicted by the model versus the actual end-of-season positions for each team, both for the full league and for home games only. The average displacement value here represents the mean deviation of the predicted positions from the actual positions, providing a measure of the model's accuracy, with a specific emphasis on the data from home games, as all models are fine-tuned using this subset of data. This analysis allows us to understand the model's predictive capability and identify where it has over or under-estimated team performances.

\begin{table}[ht]
\centering
\begin{tabular}{|c|c|c|c|c|c|}
\hline
\textbf{} & \textbf{} & \textbf{} & & \textbf{EoS} & \\
\textbf{} & \textbf{Exp.} & \textbf{EoS} & & \textbf{Home} & \\
\textbf{Team} & \textbf{Place} & \textbf{Place} & & \textbf{Place} & \\
\hline
Man City & 1 & 1 & - & 1 & - \\
Liverpool & 2 & 4 & $\downarrow$2 & 4 & $\downarrow$2 \\
Tottenham & 3 & 3 & - & 5 & $\downarrow$2 \\
Arsenal & 4 & 6 & $\downarrow$2 & 2 & $\uparrow$2 \\
Man United & 5 & 2 & $\uparrow$3 & 3 & $\uparrow$2 \\
Chelsea & 6 & 5 & $\uparrow$1 & 6 & - \\
Bournemouth & 7 & 12 & $\downarrow$5 & 15 & $\downarrow$8 \\
Huddersfield & 8 & 16 & $\downarrow$8 & 16 & $\downarrow$8 \\
Newcastle & 9 & 10 & $\downarrow$1 & 9 & - \\
Everton & 10 & 8 & $\uparrow$2 & 7 & $\uparrow$3 \\
Watford & 11 & 14 & $\downarrow$3 & 12 & $\downarrow$1 \\
Leicester & 12 & 9 & $\uparrow$3 & 10 & $\uparrow$2 \\
Southampton & 13 & 17 & $\downarrow$4 & 19 & $\downarrow$6 \\
Brighton & 14 & 15 & $\downarrow$1 & 8 & $\uparrow$6 \\
Swansea & 15 & 18 & $\downarrow$3 & 17 & $\downarrow$2 \\
West Brom & 16 & 20 & $\downarrow$4 & 20 & $\downarrow$4 \\
Burnley & 17 & 7 & $\uparrow$10 & 14 & $\uparrow$3 \\
Stoke & 18 & 19 & $\downarrow$1 & 18 & - \\
Crystal Palace & 19 & 11 & $\uparrow$8 & 13 & $\uparrow$6 \\
West Ham & 20 & 13 & $\uparrow$7 & 11 & $\uparrow$9 \\
\hline
Avg. Disp. &  & & 3,4 & & 3,3 \\
Top 6 Avg. Disp. &  & & 1,3 & & 1,3 \\
\hline
\end{tabular}
\caption{Comparing the forecasts from the finetuned LEM models against the end-of-season (EoS) tables. The average displacement (Avg. Disp.) is the average number of team positions from the actual position. We compare against the full and home tables, with the home table providing less displacement since all models are finetuned using home game data.}
\label{tab:football_positions}
\end{table}

Manchester City and Liverpool performed as expected, securing the first and second positions, respectively. While Manchester City won the Premier League with a record number of points, Liverpool went to the Champions League final, which is missing from our dataset but indicates the strength of the team. However, discrepancies arise further down the list. It is noteworthy that in the top 6, the average displacement is 1.3, while in the whole league, the displacement increases to 3.4. The reason for this is that the Premier League is extremely competitive. For example, there are cases in the 2017/18 Premier League where adding a win to a team could change the team's placement by up to 4 positions.

An in-depth analysis of the stats generated by fine-tuned models is presented in Annex \ref{annex}.

\subsection{Cristiano Ronaldo vs Lionel Messi}

\begin{figure*}[h]
  \centering
  \includegraphics[width=0.82\linewidth]{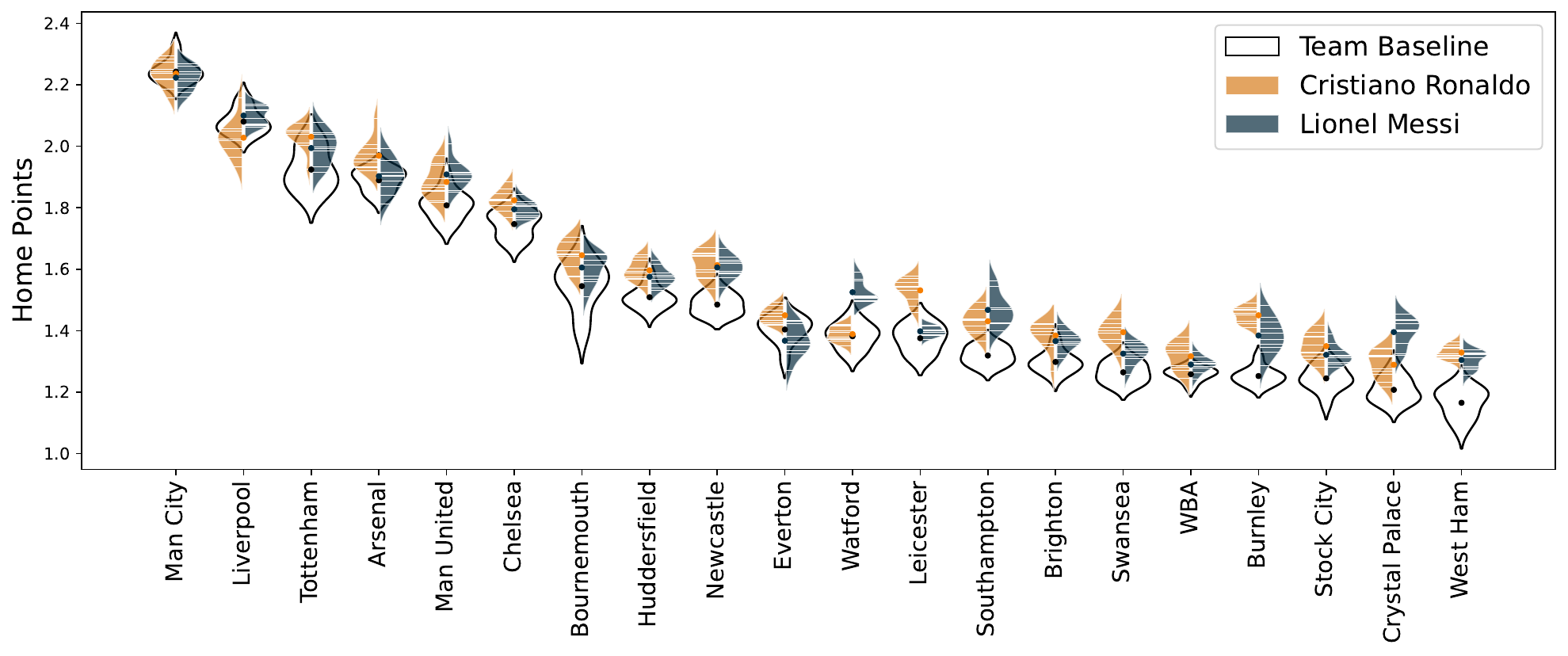}
  \caption{The expected impact of \textbf{adding} Cristiano Ronaldo or Lionel Messi on the teams in the EPL. The figure presents the violin plots of the simulations using the fine-tuned models.}
  \label{fig:finetuning_violin_player_adding}
\end{figure*}

\begin{figure*}[h]
  \centering
  \includegraphics[width=0.82\linewidth]{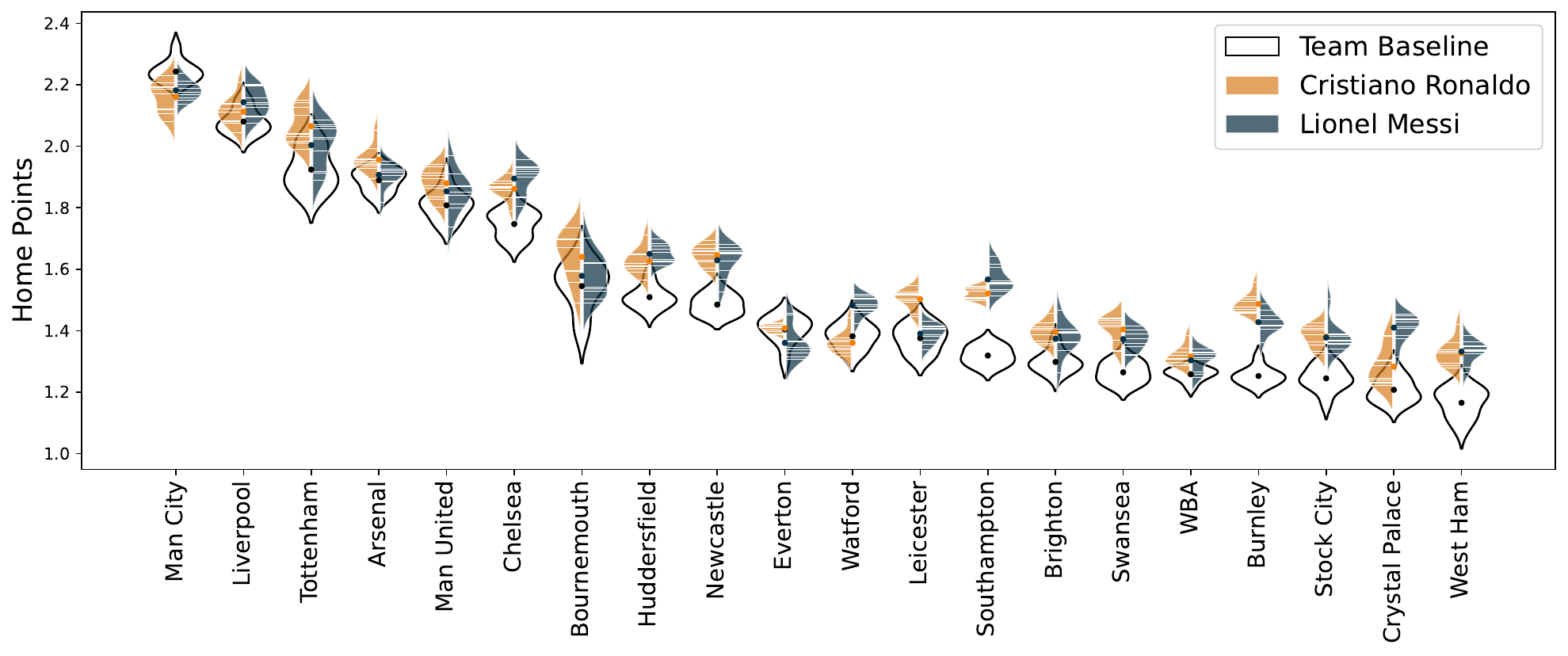}
  \caption{The expected impact of \textbf{replacing} a player for Cristiano Ronaldo or Lionel Messi on the teams in the EPL.}
  \label{fig:finetuning_violin_player_replacement}
\end{figure*}

In Figures \ref{fig:finetuning_violin_player_adding} and \ref{fig:finetuning_violin_player_replacement}, we explore the hypothetical influence of football superstars Cristiano Ronaldo and Lionel Messi on the home performance of teams. Utilizing fine-tuned predictive models, the violin plots reveal the distribution of expected home points with the addition of either player to the teams. These distributions allow us to compare the mean, variability, and shape of the teams' points distribution under the baseline conditions and the two alternative scenarios.

The "Team" data, indicated by the black and white violin, shows the distribution of points without the influence of star players. There is a noticeable variation between teams, with some, like Manchester City and Liverpool, showing a higher median and denser distribution towards the top end of the scale, suggesting consistently high performance at home games. 

The overlay in blue shows the theoretical impact of Cristiano Ronaldo on these teams. In cases such as Tottenham, there is a discernible increase in the average home points and the density towards the higher end of the points spectrum. This suggests a positive impact on the team's home performance when Ronaldo is hypothetically part of the team. The orange overlay indicates Messi's hypothetical influence. Similar to Ronaldo, Messi's presence elevates the home points. However, their influence varies across different teams, indicating that the impact is team-specific.

An interesting aspect of our results is that both players fail to improve Manchester City. We believe this is to be expected. In our dataset, Manchester City made a record-breaking season with 100 points. This represents a team executing at an exceptional level; therefore, any change would likely disrupt the team's currently optimized routines.

There is a fact that we did not analyze in-depth, but it is of crucial importance. Let us look at the cases of Watford and Leicester. Messi performs much better than Ronaldo in Watford, while the opposite happens in Leicester. The reason is largely attributed to the players already performing in the roles they would fill. Messi is a much better player when compared with Watford's Roberto Pereyra or André Carrillo than he is when compared with Leicester's Riyad Mahrez. For Ronaldo, he is a much better player than Leicester's Demarai Gray than he is when compared with Watford's Richarlison. These are the contexts that we aimed to bring with our LEMs. The current state of a team and its players has a very high impact on the outcomes measured.

The way players change the team's behavior is also very important. We present the results as distributions rather than absolute values because there are many aspects of the distribution of the results that are interesting to analyze. For example, introducing Messi or Ronaldo to the teams in the Premier League generally leads to a reduction in the variance of the distribution. This is a result since decreasing variance can be advantageous in certain situations. For example, the likelihood of relegation can be decreased by changing the variance of the team, even at the expense of decreasing the average outcome.

In player replacement analysis, the outcome suggests that we can calculate it by adding the influence of the incoming player and subtracting the influence of the outgoing player. This finding implies that we might be able to combine individual player impact distributions from separate simulations under specific conditions, which could help reduce the time spent on simulations.

\subsection{The importance of context}

\begin{figure}[h]
  \centering
  \includegraphics[width=0.63\linewidth]{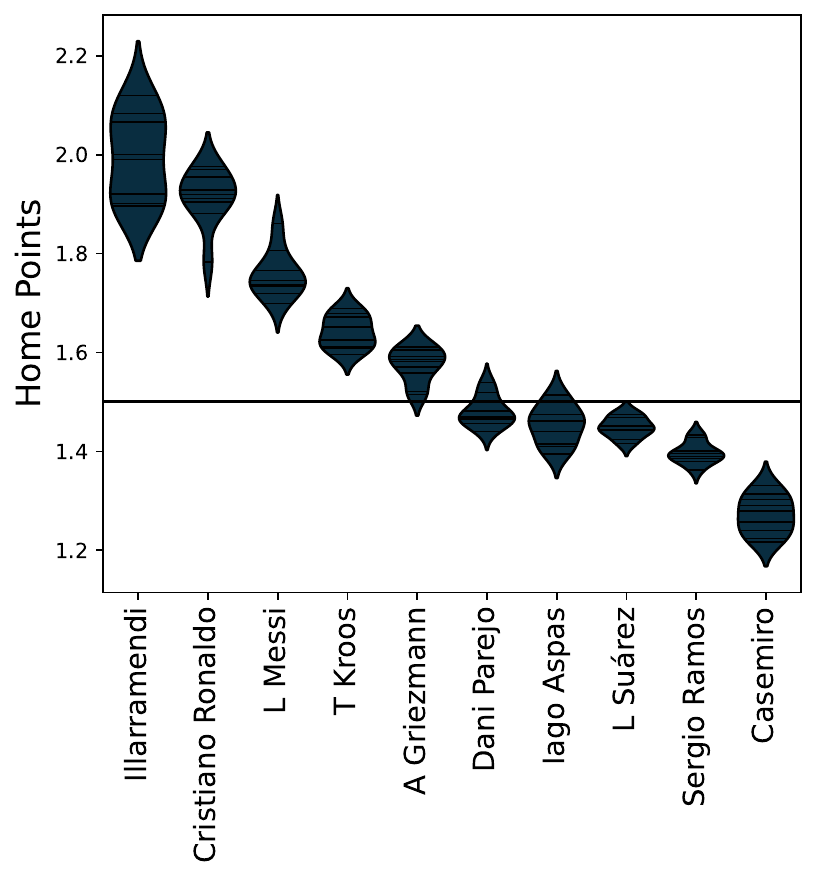}
  \caption{The impact on the baseline of the top 10 players of the 2017/2018 season, according to Sofascore.}
  \label{fig:finetuning_violin_players_all}
\end{figure}

\begin{figure*}[h]
  \centering
  \includegraphics[width=0.82\linewidth]{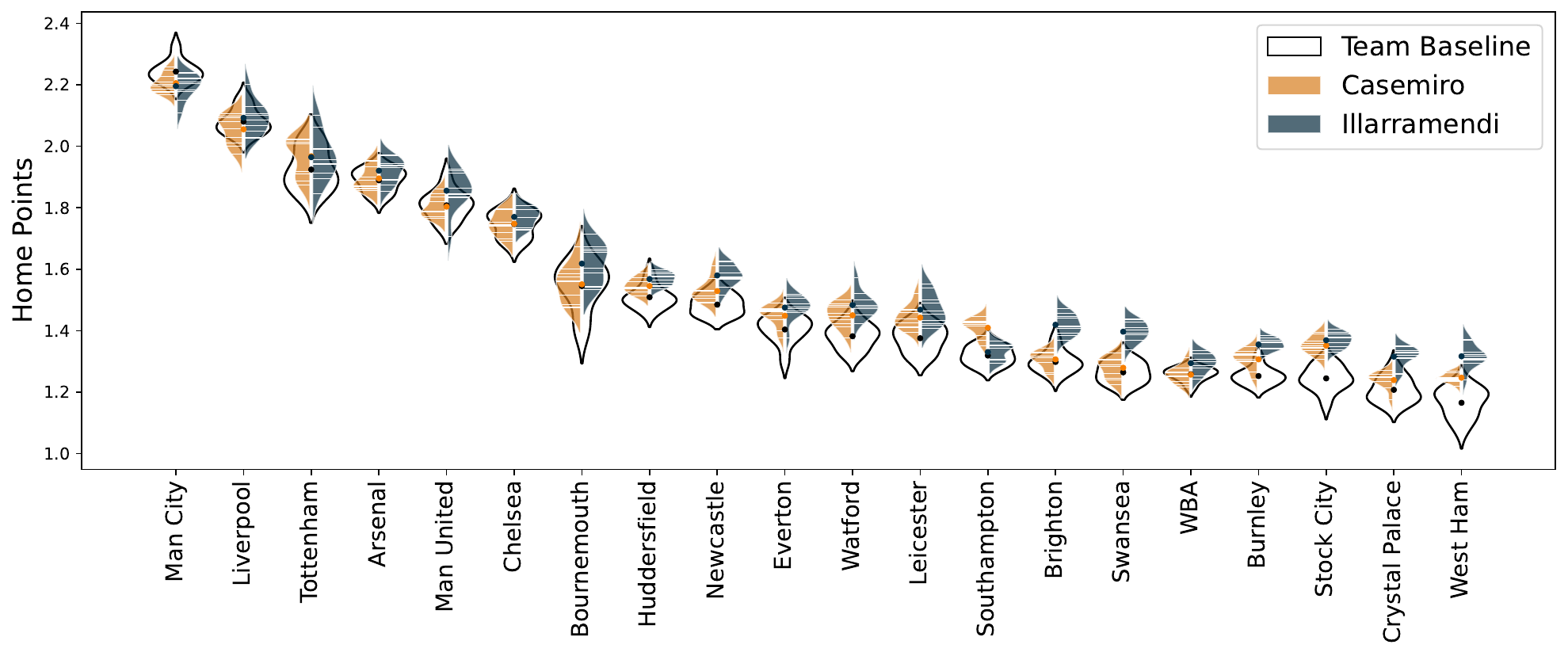}
  \caption{The expected impact of adding Casemiro or Illarramendi on the teams in the EPL.}
  \label{fig:finetuning_violin_player_adding_alt_casemiro_iral}
\end{figure*}

In Figure \ref{fig:finetuning_violin_players_all} we present the results of fine-tuning the models using only data from the players. We observe that there are several unexpected results. To explore further, we analyze the impact of adding the best and worst performers to the teams in the Premier League in Figure \ref{fig:finetuning_violin_player_adding_alt_casemiro_iral}.

Figures \ref{fig:finetuning_violin_players_all} and \ref{fig:finetuning_violin_player_adding_alt_casemiro_iral} have the role of explaining that context is everything. Figure \ref{fig:finetuning_violin_players_all} presents the expected impact of players if they were placed in the original context of the LEM, i.e., placed in a team that is the average of all teams. Figure 9 shows why it is wrong to use this general context. We observe that, when introducing the best (Illarramendi) and the worst (Casemiro) performers from Figure \ref{fig:finetuning_violin_players_all} in the teams in the Premier League, the difference between them is smaller than first suggested. While Illarramendi still outperforms Casemiro most times, the difference between the impact on the teams is lower than expected.

\section{Discussion}\label{sec:discussion}

These results show that LEMs can play a fundamental role in sports analytics, especially in evaluating players under context. LEMs can understand and learn a team's playstyle, enabling us to probe a player in much more detail and across a wide range of metrics. While this work is limited to analyzing the Points Per Game metric, many other metrics can be derived from this method. Note that LEMs generate event data; therefore, any metric based on event data can be used to evaluate the player in the new context.

Nonetheless, several challenges need to be addressed. Measuring context with event data is limited. It limits the data to the events where the player interacts with the ball. Furthermore, we cannot introduce more events, for example, passes received, because they would be introducing information that is not exclusive to the player but also from their teammates. We cannot completely isolate the player from its current team context. Therefore, the results we present are influenced in a certain amount by the current context of the player and, at the same time, are missing some crucial information about the player that is not captured in event data or, when captured, cannot be used because it contains the context of multiple players. The frontier between one player's data and another is not clear-cut but rather a fuzzy frontier.

From the practical standpoint, LEMs struggle to reason the potential changes that a player can bring to the play model of a team. For example, if Messi went to Stoke City, they would avoid playing a direct style of play. This is leading to an underestimation of the actual value that these star players would bring to a team. For teams playing a style that suits them, they provide enormous value. However, for teams whose style fits less, they provide little value. This is not a fault of the model but rather a limitation. LEMs are suitable to evaluate how a player would fit in the current context rather than in the hypothetical context that the team will play when the player arrives.

Many aspects of the approach can be improved: training LEMs with larger datasets, especially including more data from the target league, using multi-season datasets to validate the metrics obtained across different transfers, increasing the context size of the LEM, currently limited to 1 event but can be extended to a much larger number of events, and test more deep learning architectures such as using Transformers and Snapshot Ensembles to improve the model accuracy. The improvement suggestions require two types of resources: increasing the size of the datasets available and increasing the computational power. Regarding the latter, it is also important to explore potential computational efficiencies that can be used to generate faster results.

\section{Conclusion}\label{sec:conclusion}

This study has demonstrated the effectiveness and limitations of LEMs in soccer, particularly in the context of the EPL. We observed that the baseline performances of teams like Manchester City and Liverpool were highly predictable. In contrast, teams such as West Ham and Stoke City presented more variability in match outcomes.

Adding star players like Ronaldo and Messi generally led to more predictable outcomes, evidenced by a decrease in log-loss for most teams. However, their impact varied depending on team dynamics and playstyles. For instance, their inclusion did not notably improve Manchester City's performance, likely due to their already optimized play.

Our findings suggest that LEMs are a valuable tool in sports analytics, capable of understanding and learning team playstyles. This allows for a detailed analysis of players in various contexts using metrics derived from event data, allowing for insights that can influence the player recruitment strategy. However, limitations exist in the current approach, particularly in isolating player influence from team context and predicting changes a player might bring to a team's playstyle.

While LEMs show great potential in player and team performance analysis, their effectiveness is contingent on the context and dynamics of the team and league. As such, they represent an evolving tool in sports analytics, with scope for significant advancements and broader applications in the future.

\begin{ack}
This work is financed by National Funds through the Portuguese funding agency, FCT - Fundação para a Ciência e a Tecnologia, within project UIDB/50014/2020.
\end{ack}

\bibliography{mybibfile}

\begin{thebibliography}{18}
\providecommand{\natexlab}[1]{#1}
\providecommand{\url}[1]{\texttt{#1}}
\expandafter\ifx\csname urlstyle\endcsname\relax
  \providecommand{\doi}[1]{doi: #1}\else
  \providecommand{\doi}{doi: \begingroup \urlstyle{rm}\Url}\fi

\bibitem[Decroos et~al.(2019)Decroos, Bransen, Van~Haaren, and Davis]{decroos_actions_2019}
T.~Decroos, L.~Bransen, J.~Van~Haaren, and J.~Davis.
\newblock Actions {Speak} {Louder} than {Goals}: {Valuing} {Player} {Actions} in {Soccer}.
\newblock In \emph{Proceedings of the 25th {ACM} {SIGKDD} {International} {Conference} on {Knowledge} {Discovery} \& {Data} {Mining}}, pages 1851--1861, Anchorage AK USA, July 2019. ACM.
\newblock ISBN 978-1-4503-6201-6.
\newblock \doi{10.1145/3292500.3330758}.
\newblock URL \url{https://dl.acm.org/doi/10.1145/3292500.3330758}.

\bibitem[Fernández et~al.(2021)Fernández, Bornn, and Cervone]{fernandez_framework_2021}
J.~Fernández, L.~Bornn, and D.~Cervone.
\newblock A framework for the fine-grained evaluation of the instantaneous expected value of soccer possessions.
\newblock \emph{Machine Learning}, 110\penalty0 (6):\penalty0 1389--1427, June 2021.
\newblock ISSN 0885-6125, 1573-0565.
\newblock \doi{10.1007/s10994-021-05989-6}.
\newblock URL \url{https://link.springer.com/10.1007/s10994-021-05989-6}.

\bibitem[Hakes and Sauer(2006)]{hakes_economic}
J.~K. Hakes and R.~D. Sauer.
\newblock An economic evaluation of the moneyball hypothesis.
\newblock \emph{Journal of Economic Perspectives}, 20\penalty0 (3):\penalty0 173--186, September 2006.
\newblock \doi{10.1257/jep.20.3.173}.
\newblock URL \url{https://www.aeaweb.org/articles?id=10.1257/jep.20.3.173}.

\bibitem[Huang and Chang(2023)]{huang2023reasoning}
J.~Huang and K.~C.-C. Chang.
\newblock Towards reasoning in large language models: A survey, 2023.

\bibitem[Lewis(2004)]{lewis_moneyball_2004}
M.~Lewis.
\newblock \emph{Moneyball: {The} {Art} of {Winning} an {Unfair} {Game}}.
\newblock Business book summary. W. W. Norton, 2004.
\newblock ISBN 978-0-393-32481-5.
\newblock URL \url{https://books.google.pt/books?id=3DnUCWSnci0C}.

\bibitem[McHale and Holmes(2023)]{mchale_estimating_2023}
I.~G. McHale and B.~Holmes.
\newblock Estimating transfer fees of professional footballers using advanced performance metrics and machine learning.
\newblock \emph{European Journal of Operational Research}, 306\penalty0 (1):\penalty0 389--399, Apr. 2023.
\newblock ISSN 03772217.
\newblock \doi{10.1016/j.ejor.2022.06.033}.
\newblock URL \url{https://linkinghub.elsevier.com/retrieve/pii/S0377221722005082}.

\bibitem[Mendes-Neves et~al.(2021)Mendes-Neves, Mendes-Moreira, and Rossetti]{mendes_neves_datadriven_simulator_2021}
T.~Mendes-Neves, J.~Mendes-Moreira, and R.~J.~F. Rossetti.
\newblock A {Data}-{Driven} {Simulator} for {Assessing} {Decision}-{Making} in {Soccer}.
\newblock In G.~Marreiros, F.~S. Melo, N.~Lau, H.~Lopes~Cardoso, and L.~P. Reis, editors, \emph{Progress in {Artificial} {Intelligence}}, volume 12981, pages 687--698. Springer International Publishing, Cham, 2021.
\newblock ISBN 978-3-030-86229-9 978-3-030-86230-5.
\newblock \doi{10.1007/978-3-030-86230-5_54}.
\newblock URL \url{https://link.springer.com/10.1007/978-3-030-86230-5_54}.
\newblock Series Title: Lecture Notes in Computer Science.

\bibitem[Mendes-Neves et~al.(2022)Mendes-Neves, Meireles, and Mendes-Moreira]{mendesneves2022valuing}
T.~Mendes-Neves, L.~Meireles, and J.~Mendes-Moreira.
\newblock Valuing players over time, 2022.

\bibitem[Mendes-Neves et~al.(2024)Mendes-Neves, Meireles, and Mendes-Moreira]{mendes_neves_towards_LEM}
T.~Mendes-Neves, L.~Meireles, and J.~Mendes-Moreira.
\newblock Towards a foundation large events model for soccer.
\newblock \emph{[Manuscript under revision at the Machine Learning Journal]}, 2024.

\bibitem[Pappalardo et~al.(2019)Pappalardo, Cintia, Rossi, Massucco, Ferragina, Pedreschi, and Giannotti]{pappalardo_public_2019}
L.~Pappalardo, P.~Cintia, A.~Rossi, E.~Massucco, P.~Ferragina, D.~Pedreschi, and F.~Giannotti.
\newblock A public data set of spatio-temporal match events in soccer competitions.
\newblock \emph{Scientific Data}, 6\penalty0 (1):\penalty0 236, Dec. 2019.
\newblock ISSN 2052-4463.
\newblock \doi{10.1038/s41597-019-0247-7}.
\newblock URL \url{http://www.nature.com/articles/s41597-019-0247-7}.

\bibitem[Pollard et~al.(2004)Pollard, Ensum, and Taylor]{Pollard_2004}
R.~Pollard, J.~Ensum, and S.~Taylor.
\newblock Estimating the probability of a shot resulting in a goal: The effects of distance, angle and space.
\newblock \emph{Int. J. Soccer Sci.}, 2, 01 2004.

\bibitem[Radford et~al.(2018)Radford, Narasimhan, Salimans, and Sutskever]{radford_improving_nodate}
A.~Radford, K.~Narasimhan, T.~Salimans, and I.~Sutskever.
\newblock Improving {Language} {Understanding} by {Generative} {Pre}-{Training}.
\newblock 2018.
\newblock URL \url{https://cdn.openai.com/research-covers/language-unsupervised/language_understanding_paper.pdf}.

\bibitem[Simpson et~al.(2022)Simpson, Beal, Locke, and Norman]{simpson_seq2event_2022}
I.~Simpson, R.~J. Beal, D.~Locke, and T.~J. Norman.
\newblock {Seq2Event}: {Learning} the {Language} of {Soccer} {Using} {Transformer}-based {Match} {Event} {Prediction}.
\newblock In \emph{Proceedings of the 28th {ACM} {SIGKDD} {Conference} on {Knowledge} {Discovery} and {Data} {Mining}}, pages 3898--3908, Washington DC USA, Aug. 2022. ACM.
\newblock ISBN 978-1-4503-9385-0.
\newblock \doi{10.1145/3534678.3539138}.
\newblock URL \url{https://dl.acm.org/doi/10.1145/3534678.3539138}.

\bibitem[{Statsbomb}()]{statsbomb_what_nodate}
{Statsbomb}.
\newblock What is {Expected} {Threat} ({xT})? {Possession} {Value} models explained.
\newblock URL \url{https://statsbomb.com/soccer-metrics/possession-value-models-explained/}.

\bibitem[{Tahmeed Tureen} and {SBH Olthof}(2022)]{tahmeed_tureen_estimated_2022}
{Tahmeed Tureen} and {SBH Olthof}.
\newblock “{Estimated} {Player} {Impact}” ({EPI}): {Quantifying} the effects of individual players on football (soccer) actions using hierarchical statistical models.
\newblock In \emph{{StatsBomb} {Conference} {Proceedings}}, Wembley, London, 2022.

\bibitem[{The Come On Man}()]{the_come_on_man_predicting_nodate}
{The Come On Man}.
\newblock Predicting the {Success} of a {Transfer}.
\newblock URL \url{https://thecomeonman.github.io/PredictingTransferSuccesses/}.

\bibitem[{Tom Worville}(2021)]{tom_worville_how_2021}
{Tom Worville}.
\newblock How football’s finest are using analytics to find an edge, 2021.
\newblock URL \url{https://theathletic.com/2882187/2021/10/12/how-to-find-the-edge-examining-premier-league-analytics-trends-what-is-to-come/}.

\bibitem[Yeung et~al.(2023)Yeung, Sit, and Fujii]{yeung_transformer-based_2023}
C.~C.~K. Yeung, T.~Sit, and K.~Fujii.
\newblock Transformer-{Based} {Neural} {Marked} {Spatio} {Temporal} {Point} {Process} {Model} for {Football} {Match} {Events} {Analysis}, Feb. 2023.
\newblock URL \url{http://arxiv.org/abs/2302.09276}.
\newblock arXiv:2302.09276 [cs].

\end{thebibliography}

\clearpage
\appendix

\section{Stats from Simulations} \label{annex}

\begin{table*}[h!]
\centering
\begin{tabular}{|c|c|c|c|c|c|c|c|c|c|c|c|c|c|c|}
\hline
               &                 &                 & \textbf{Attack} & \textbf{Attack} & \textbf{Def.}  & \textbf{Def.}  & \textbf{Aerial} & \textbf{Aerial} &                &                &                &                 \\
               & \textbf{Passes} & \textbf{Passes} & \textbf{Duels}  & \textbf{Duels}  & \textbf{Duels} & \textbf{Duels} & \textbf{Duels}  & \textbf{Duels}  & \textbf{Shots} & \textbf{Shots} & \textbf{Goals} & \textbf{Goals}  \\
\textbf{Team}  & \textbf{Home}   & \textbf{Away}   & \textbf{Home}   & \textbf{Away}   & \textbf{Home}  & \textbf{Away}  & \textbf{Home}   & \textbf{Away}   & \textbf{Home}  & \textbf{Away}  & \textbf{Home}  & \textbf{Away}   \\ 
\hline
Man City       & 621             & 283             & 72              & 53              & 54             & 71             & 27              & 27              & 16             & 7              & 2.8            & 1.2             \\
Liverpool      & 516             & 343             & 78              & 61              & 62             & 77             & 39              & 38              & 16             & 8              & 2              & 0.9             \\
Tottenham      & 490             & 320             & 76              & 63              & 65             & 74             & 40              & 41              & 15             & 10             & 1.7            & 0.8             \\
Arsenal        & 509             & 353             & 69              & 67              & 68             & 69             & 41              & 41              & 15             & 10             & 2.2            & 1.4             \\
Man United     & 459             & 371             & 76              & 67              & 67             & 76             & 35              & 37              & 12             & 9              & 1.4            & 0.9             \\
Chelsea        & 467             & 388             & 77              & 67              & 68             & 76             & 37              & 38              & 15             & 10             & 1.3            & 0.9             \\
Bournemouth    & 379             & 391             & 70              & 68              & 69             & 70             & 47              & 48              & 12             & 11             & 1.3            & 1.2             \\
Huddersfield   & 369             & 395             & 69              & 71              & 71             & 69             & 49              & 50              & 9              & 9              & 1              & 0.8             \\
Newcastle      & 337             & 398             & 65              & 71              & 71             & 66             & 57              & 56              & 11             & 11             & 1              & 0.8             \\
Everton        & 344             & 375             & 68              & 72              & 72             & 68             & 58              & 55              & 9              & 11             & 1.1            & 1               \\
Watford        & 370             & 348             & 73              & 71              & 71             & 73             & 53              & 54              & 11             & 10             & 1.2            & 1.2             \\
Leicester      & 365             & 361             & 72              & 71              & 71             & 72             & 48              & 48              & 10             & 10             & 1.1            & 1               \\
Southampton    & 405             & 355             & 70              & 66              & 66             & 69             & 44              & 44              & 12             & 11             & 1              & 1.1             \\
Brighton       & 365             & 421             & 62              & 67              & 67             & 62             & 49              & 51              & 10             & 11             & 0.9            & 1               \\
Swansea        & 383             & 427             & 66              & 69              & 69             & 66             & 46              & 47              & 10             & 11             & 1              & 1.1             \\
WBA            & 325             & 409             & 67              & 70              & 70             & 66             & 51              & 51              & 10             & 11             & 1.2            & 1.3             \\
Burnley        & 331             & 363             & 60              & 62              & 62             & 61             & 59              & 61              & 10             & 11             & 0.7            & 0.7             \\
Stoke City     & 305             & 384             & 64              & 68              & 68             & 64             & 57              & 54              & 10             & 12             & 1.1            & 1.3             \\
Crystal Palace & 343             & 385             & 76              & 77              & 76             & 76             & 44              & 46              & 11             & 11             & 1.1            & 1.3             \\
West Ham       & 348             & 375             & 73              & 69              & 69             & 73             & 47              & 51              & 10             & 10             & 1              & 1.4             \\
\hline
\end{tabular}
\caption{Per game average of several team statistics in our simulations. The 'Home' columns detail the actions performed by the team, while the 'Away' columns represent the actions performed by the opposition teams}
\label{tab:football_statistics}
\end{table*}

The accompanying Table \ref{tab:football_statistics} delves into a comprehensive set of per-game statistics from our simulations, segmented into actions performed by the team (home) and by the opponent (away). This breakdown provides insights into the teams' performances in different settings. Key metrics include the number of passes, attacking, defensive, aerial duels, shots, and goals. Such a detailed statistical analysis is crucial for understanding the dynamics of each team's style and strategies.

We see clear evidence that the LEM can learn a team's playstyle and patterns. Manchester City has the highest number of passes per game, indicating a playing style emphasizing possession. At the same time, teams like Burnley and Stoke City have lower pass counts, suggesting a more direct style of play during games. Furthermore, the latter teams are among the highest-ranked teams in terms of Aerial Duels, emphasizing the direct style of the team. A high number of passes with a corresponding high number of shots and goals suggests a possession-based approach that effectively creates and utilizes scoring opportunities. In contrast, teams with lower pass counts but higher aerial duels and crosses might prioritize physicality and set pieces as key elements of their game plan.

\end{document}